\documentclass{article}

\PassOptionsToPackage{numbers, compress, square}{natbib}

\usepackage[final]{neurips_2019}

\usepackage[utf8]{inputenc} 
\usepackage[T1]{fontenc}    
\usepackage{hyperref}       
\usepackage{url}            
\usepackage{booktabs}       
\usepackage{amsfonts}       
\usepackage{nicefrac}       
\usepackage{microtype}      
\usepackage{gensymb}
\usepackage[caption=false,font=footnotesize]{subfig}
\usepackage{float}
\usepackage{graphicx}
\usepackage{fixltx2e}
\usepackage{hyperref}
\title{Quantifying Urban Canopy Cover with Deep Convolutional Neural Networks}

\author{%
  Bill~Cai\thanks{Also affiliated with the Government Technology Agency of Singapore, Singapore 117438} \\
  Center for Computational Engr.\\
  MIT\\
  Cambridge, MA 02139 \\
  \texttt{billcai@alum.mit.edu} \\  
  \And
	Xiaojiang~Li \\
	Dept. of Geog. and Urban Studies\\
	Temple University\\
	Philadelphia, PA 19122 \\
	\texttt{xiaojian@mit.edu} \\
\And
	Carlo~Ratti \\
	Senseable City Lab\\
	MIT\\
	Cambridge, MA 02139 \\
	\texttt{ratti@mit.edu}
}
\begin{document}

\maketitle

\begin{abstract}
Urban canopy cover is important to mitigate the impact of climate change. Yet, existing quantification of urban greenery is either manual and not scalable, or use traditional computer vision methods that are inaccurate. We train deep convolutional neural networks (DCNNs) on datasets used for self-driving cars to estimate urban greenery instead, and find that our semantic segmentation and direct end-to-end estimation method are more accurate and scalable, reducing mean absolute error of estimating the Green View Index (GVI) metric from 10.1\% to 4.67\%. With the revised DCNN methods, the \textit{Treepedia} project was able to scale and analyze canopy cover in 22 cities internationally, sparking interest and action in public policy and research fields.
\end{abstract}

\section{Introduction}

Urban canopy cover is generally acknowledged as an effective way of mitigating the impact of increasing daytime summer temperatures \cite{norton2015planning,bowler2010urban} that have recently reached record highs internationally \cite{jonhenley,copernicus2019}. Physical models show that urban trees can significantly reduce the diurnal temperature range \cite{kusaka2004coupling}, while empirical studies demonstrate that urban canopy cover have reduced peak summer air temperatures by up to 2.8\degree C, 1.5\degree C, 2.0 \degree C, and 2.7 \degree C in Campinas (Brazil), Singapore, Shanghai (China), and Freiburg (Germany) respectively \cite{de2015effect,wong2010study,yang2011thermal,lee2016contribution}. Existing studies have established a direct relationship between high peak summer temperatures and critical health outcomes, such as hospital admissions \cite{chan2013hospital,michelozzi2009high}, deaths \cite{michelozzi2005impact,fouillet2008has,semenza1996heat}, respiratory diseases and cardiovascular health \cite{lin2009extreme,uejio2016summer}. High temperatures can also significantly depress the economic growth of developing \cite{dell2012temperature} and developed countries \cite{colacito2016temperature} by decreasing labor productivity and supply \cite{graff2014temperature}, and increasing political instability \cite{dell2012temperature,burke2015climate}.

Besides decreasing air temperature in cities, benefits of trees and canopy cover in urban areas also include removal of air pollution \cite{nowak2006air}, increased perceived neighborhood safety \cite{li2015does}, and better visual and aesthetic appeal for residents \cite{jackson2003relationship,lindemann2016does}.

\textbf{Challenges of quantifying canopy cover:} Current methods to measure existing urban canopy cover remain inadequate. Traditional methods rely on either overhead imagery or in-person fieldwork. High resolution overhead images are often costly to obtain, hence limiting most analysis to coarse resolutions \cite{mathieu2007mapping,thomas2003comparison}. Overhead imagery also cannot represent the street-level and resident perspectives of canopy cover \cite{yang2009can}. In-person fieldwork requires significant man-hours to cover large urban areas \cite{treesingapore}.

First proposed by \citet{yang2009can} and later used by \citet{li2015assessing}, the Green View Index (GVI) measures the street-level urban canopy cover by averaging percentage of canopy cover on a pixel-level in street level images. Existing methods \cite{li2015assessing,yu2019bgvi} that calculate GVI primarily relies on the original method of color thresholding and clustering to filter for possibly misidentified green specks. The GVI metric has proven to be applicable to other sources of street-level imagery, including Tencent Street View \cite{long2017green,dong2018green} and Baidu Street View \cite{yu2019bgvi}, and also resulted in findings of relationships between the prevalence of urban greenery to neighborhood wealth \cite{li2015lives}, property prices \cite{zhang2018impacts}, cycling and walking behaviors \cite{lu2018effect}. These subsequently discovered relationships underline the value of accurate analysis of urban canopy cover at large scales. Conversely, insufficient quantification of urban trees will decrease the effectiveness and fairness of publicly-funded urban greenery efforts.

As demonstrated in Figure 3, existing "threshold and cluster" methods are prone to false positive detection of green objects that are not considered as vertical vegetation, or false negative detection of non-green parts of vertical vegetation such as branches and yellow leaves.\footnotetext{Calculated using "threshold and cluster" method provided for by authors of \citet{li2015assessing} in \url{https://github.com/mittrees/Treepedia_Public}} Existing methods also validate their accuracy by calculating the Pearson's correlation coefficient between manually labeled and calculated GVIs \cite{li2015assessing,yu2019bgvi,long2017green,dong2018green}. The Pearson's correlation coefficient measures the strength of co-movements between manually labeled and calculated GVI values, but do not provide a direct measurement of difference between manually labeled and calculated GVI values and also do not capture the accuracy of located pixels of vertical vegetation \cite{cai2018treepedia}.

\section{Dataset and Methods}

\textbf{Dataset:} We choose Cambridge (USA), Johannesburg (South Africa), Oslo (Norway), S$\widetilde{\mbox{a}}$o Paulo (Brazil), and Singapore (Singapore) as cities included in our training and test sets. From each of the 5 cities, we randomly select 100 available Google Street View (GSV) images along street networks to form a training-validation-test set. We then divide the 500 image dataset into a 100 image test set, 320 image training set and a 80 image validation set. We produce manual labels by carefully tracing all vertical vegetation in the images for all 500 images.

We augment our model training by first using the Cityscapes dataset to train our DCNN model. Designed with the use case of autonomous driving in mind, the finely labelled subset of the Cityscapes dataset contains over 5000 images taken from vehicle-mounted cameras placed in vehicles and sampled across numerous German cities \cite{cordts2016cityscapes}. We convert the Cityscapes dataset by collapsing the original multi-class labels into binary labels for vegetation and non-vegetation. By first training our models on the larger Cityscapes dataset, we increase our training dataset with the aim of increasing our model performance. 

\textbf{Metrics:} In addition to Pearson's correlation coefficient, we propose two evaluation metrics to compare tree cover estimation: mean Intersection-over-Union (mean IoU) for measuring the accuracy of the location of labelled vegetation labels, and Mean Absolute Error (MAE) for measuring the accuracy in estimating overall GVI.

\textbf{DCNN semantic segmentation:} We adopt Zhao. et al's \cite{zhao2017pyramid} Pyramid Scene Parsing Network (PSPNet)'s architecture to train a DCNN to segment vertical vegetation pixel labels. We first use pre-trained weights from the original PSPNet trained on the original Cityscapes datasets with its 19 class labels. We then pre-train the network again on the aforementioned transformed Cityscapes dataset with binary labels that classify vertical vegetation. Finally, we train the network on the small labelled GSV dataset itself. We show the qualitative results of the DCNN segmentation model in Figure 1.

\textbf{DCNN end-to-end learning:} We directly estimate GVI with a DCNN model. To conduct end-to-end direct learning of a single GVI value for each image, we adapt He et al's \cite{he2016deep} deep residual network (ResNet) architecture. We first initialize the network with weights that have been pretrained on the ImageNet dataset. We then pre-train the modified ResNet with the transformed Cityscapes dataset and associated true GVI labels, before training on the small labelled GSV dataset. 

The lack of an intermediate image segmentation mask makes it difficult to confirm or communicate the features that the DCNN end-to-end model has learned in order to estimate GVI. Selvaraju et al \cite{selvaraju2016grad} developed Gradient Class Activation Map (Grad-CAM) in order to produce visual explanations for learned features in convolutional layers. We apply Grad-CAM to our DCNN end-to-end model to understand whether our model has learned generalizable features. The qualitative results of applying Grad-CAM to our trained DCNN end-to-end model in Figure 1. We provide more examples of Grad-CAM applied to our trained DCNN end-to-end model in Figure 5.

\begin{figure}[H]
	\centering
	\resizebox{0.4\columnwidth}{!}{%
		\includegraphics[scale=0.5]{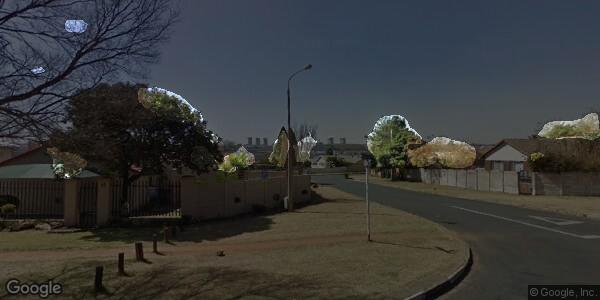}}
	\resizebox{0.4\columnwidth}{!}{%
		\includegraphics[scale=0.5]{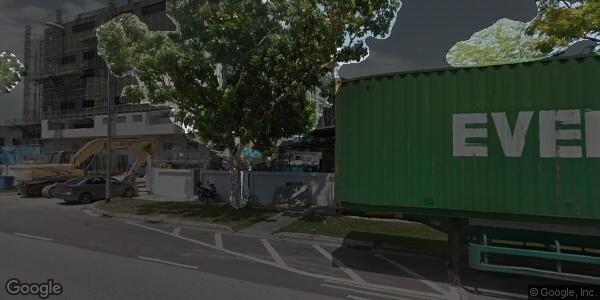}}\\
	\vspace{1mm}
	\resizebox{0.4\columnwidth}{!}{%
		\includegraphics[scale=0.5]{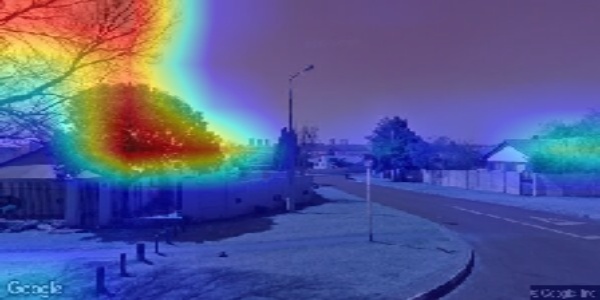}}
	\resizebox{0.4\columnwidth}{!}{%
		\includegraphics[scale=0.5]{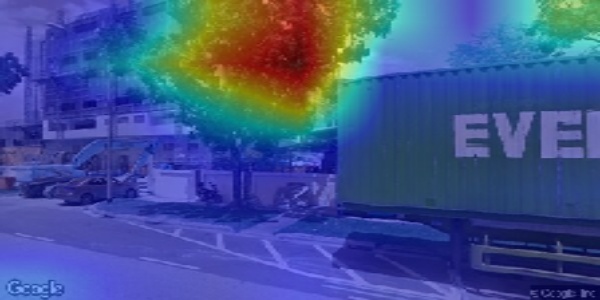}}
	\caption[]{Top: Classification of vertical vegetation by DCNN semantic segmentation model. Lighter masked areas are classified as vertical vegetation. Bottom: Grad-CAM results applied on the trained DCNN end-to-end model. Areas closer to red have a more positive contribution to a higher prediction of GVI than the contribution of areas closer to blue.}
\end{figure}

\section{Results and Discussion}

\begin{table*}[h]
	\centering
	
	\resizebox{\columnwidth}{!}{
	\begin{tabular}{|c|cccc|c|}
		\hline
		Model & Mean & Mean  & Pearson's & 5\%-95\% of & Running Time \\
		& IoU & Absolute & Correlation & GVI Estimation & for 10000\\
		& (\%) & Error (\%) & Coefficient & Error (\%) & images (seconds) \\
		\hline
		\citet{li2015assessing} "threshold & 44.7 & 10.1 & 0.708 & -26.6, 18.7 & 3665\\
		and cluster" & & && &\\
		DCNN semantic & 61.3 & 7.83 &0.830& -20.0, 12,37 & 2064\\
		segmentation & & & & &\\
		DCNN end-to-end & NA & 4.67 & 0.939&-10.9, 7.97 & 38.9\\
		\hline
	\end{tabular}}
	\caption[]{Accuracy and processing speed\footnotemark comparison between models. DCNN end-to-end model does not provide an intermediate image segmentation, hence mean IoU metrics are not computed.}
\end{table*}

DCNN models perform better than the "threshold and cluster" method across the original Pearson's correlation coefficient and metrics proposed by this paper. Furthermore, we show that the GVI estimation error 5\%-95\% bounds are substantially narrower using DCNN models.  The DCNN end-to-end model also allows for efficient quantification of GVI. To put this in perspective, the DCNN end-to-end model can process 1 million Google Street View images required to analyze urban greenery in a large city like London in an hour worth's of running time on a single benchmark machine, whereas the "threshold and cluster" evaluated will require around 4 days.\footnotetext{Processing speed comparisons were were conducted on a system equipped with an Intel i7-7700K processor, one NVIDIA GTX 1080Ti GPU, and 32GB of memory} The code and specialized Google Street View dataset are provided online\footnote{\url{https://github.com/billcai/treepedia_dl_public}} for reproducibility and for collaborating public sector agencies to implement.

These improvements in scalability and accuracy of quantifying urban canopy cover already have began to influence public greening policies. With this revised methodology, the \textit{Treepedia} project\footnote{The results of the \textit{Treepedia} project can be viewed from: \url{http://senseable.mit.edu/treepedia}} quantified urban canopy cover across 22 cities internationally. Besides widespread and global news coverage\cite{wsjtreepedia,timetreepedia,guardiantreepedia,wiredtreepedia}, the impact of \textit{Treepedia} project is attested to by public attention from policymakers from Singapore \cite{facebooktreepedia}, Paris \cite{timetreepedia}, the US Government's Climate Resilience Toolkit \cite{ustreepedia} and the World Economic Forum \cite{weftreepedia}. 

\subsubsection*{Acknowledgments}

The authors thank all members of the Senseable City Lab Consortium for sponsoring the research of the Senseable City Lab.
\small
\bibliography{references}

\newpage
\subsection*{Appendix: Treepedia visualization}
\begin{figure}[H]
	\centering
	\resizebox{\columnwidth}{!}{%
		\includegraphics[scale=0.5]{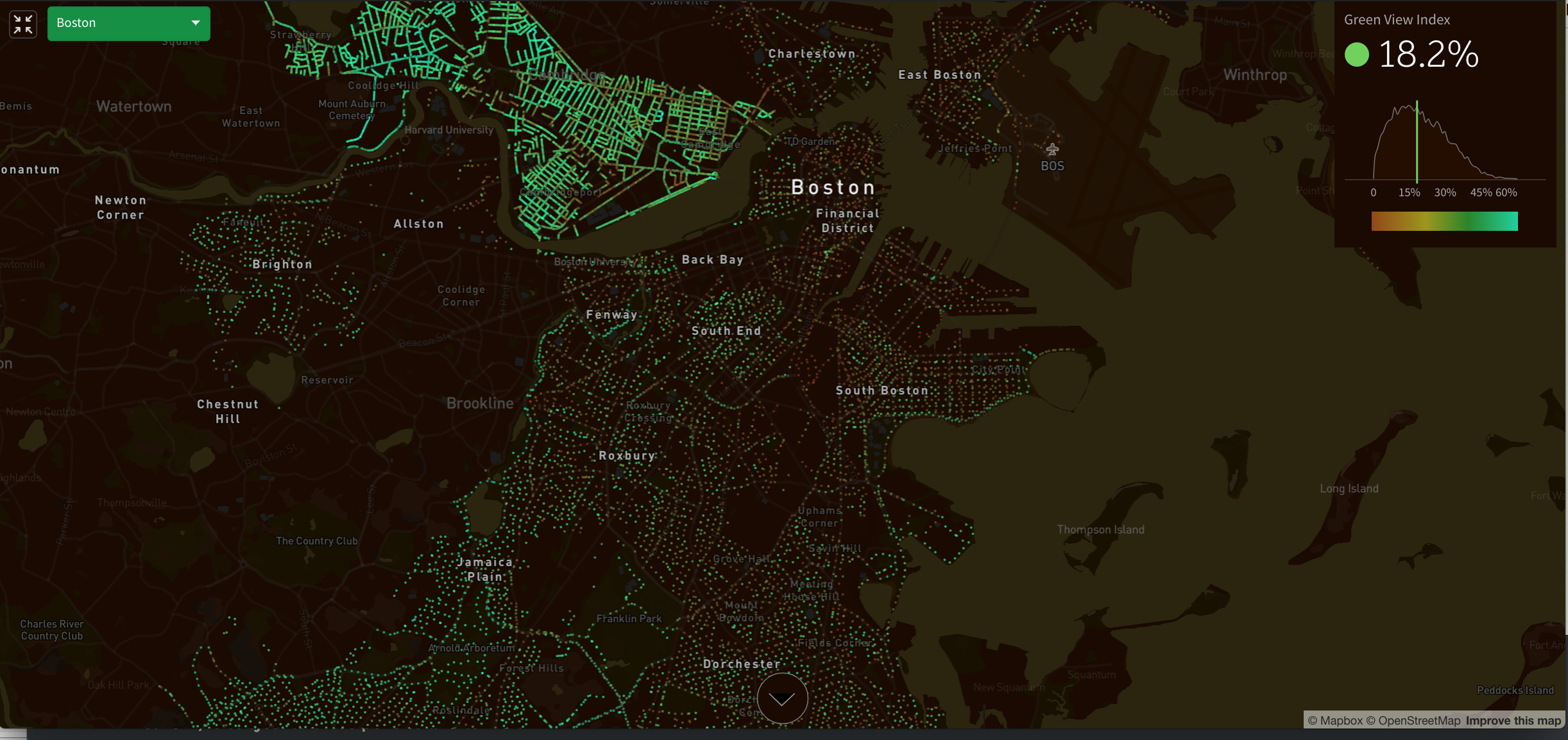}}
	\resizebox{\columnwidth}{!}{%
		\includegraphics[scale=0.5]{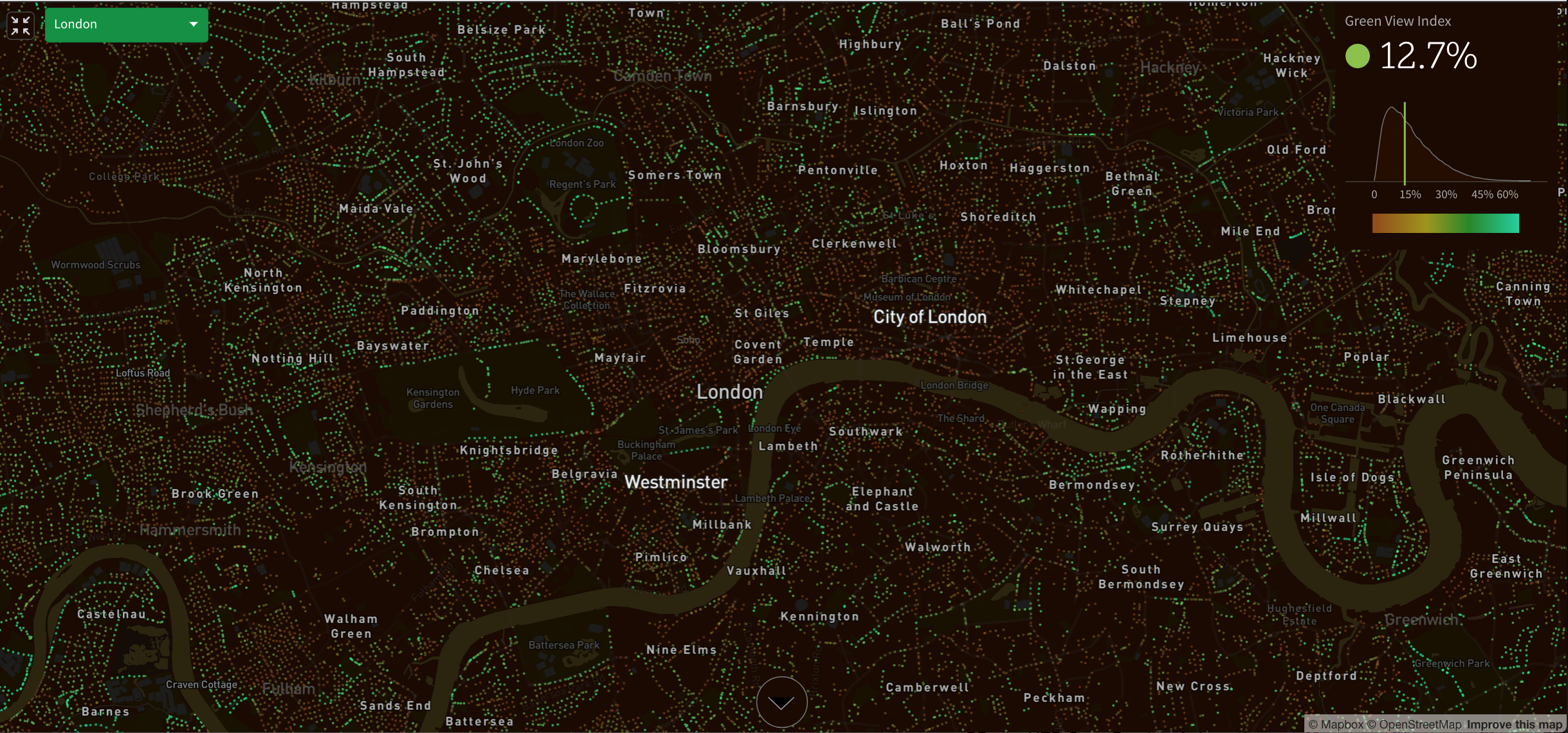}}
	\caption[]{The \textit{Treepedia} project conducted large-scale analysis of urban greenery across 22 cities. Top: Visualization of large-scale quantification of urban greenery in Boston, Bottom: Visualization of large-scale quantification of urban greenery in London}
\end{figure}

\subsection*{Appendix: Common errors of "threshold and cluster" method}

\begin{figure}[H]
	\centering
	\resizebox{0.48\columnwidth}{!}{%
		\includegraphics[scale=0.5]{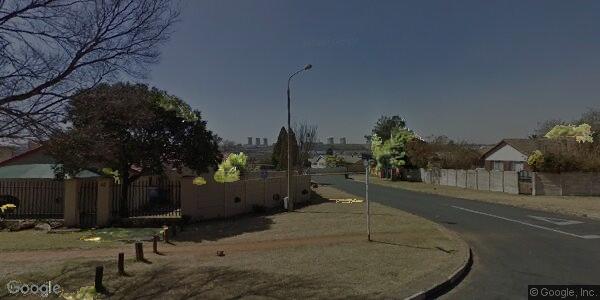}}
	\resizebox{0.48\columnwidth}{!}{%
		\includegraphics[scale=0.5]{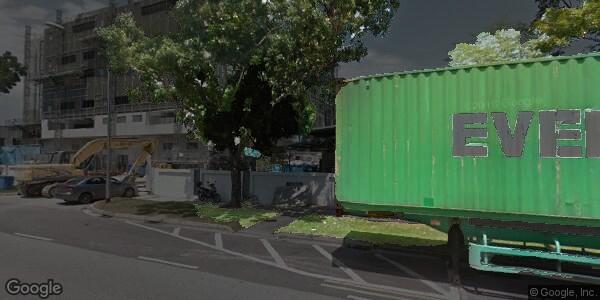}}
	\caption[]{Misclassification of vertical vegetation in test images using "threshold and cluster" method\footnotemark. Lighter masked areas are classified as vertical vegetation.}
\end{figure}

\subsection*{Appendix: Dataset visualization}

\begin{figure}[H]
	\centering
	\resizebox{0.245\columnwidth}{!}{%
		\includegraphics[scale=0.5]{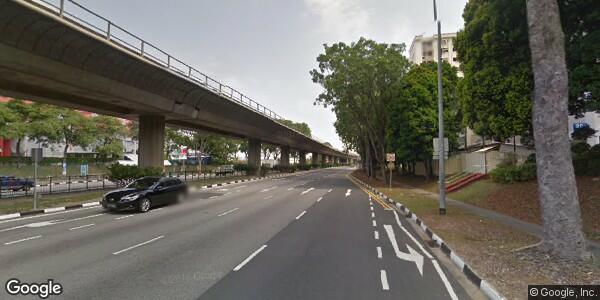}}
	\resizebox{0.245\columnwidth}{!}{%
		\includegraphics[scale=0.5]{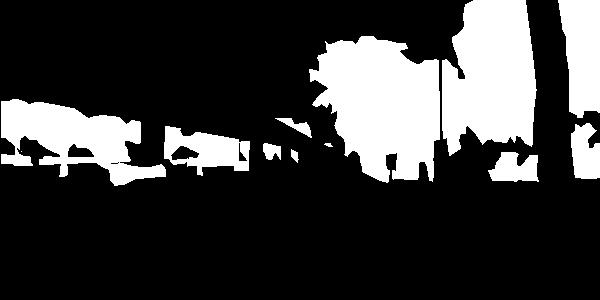}}
	\resizebox{0.245\columnwidth}{!}{%
		\includegraphics[scale=0.5]{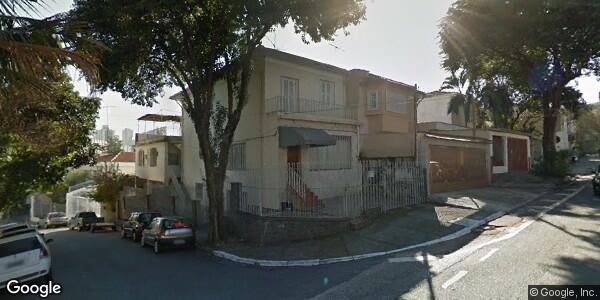}}
	\resizebox{0.245\columnwidth}{!}{%
		\includegraphics[scale=0.5]{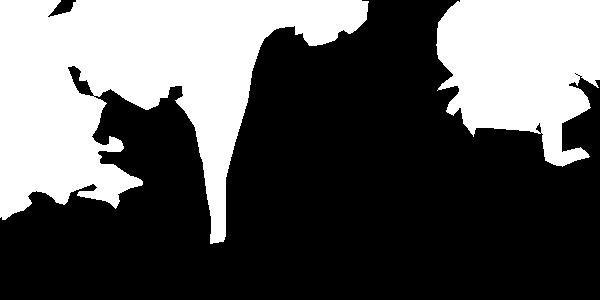}}\\
	\vspace{1mm}
	\resizebox{0.245\columnwidth}{!}{%
		\includegraphics[scale=0.5]{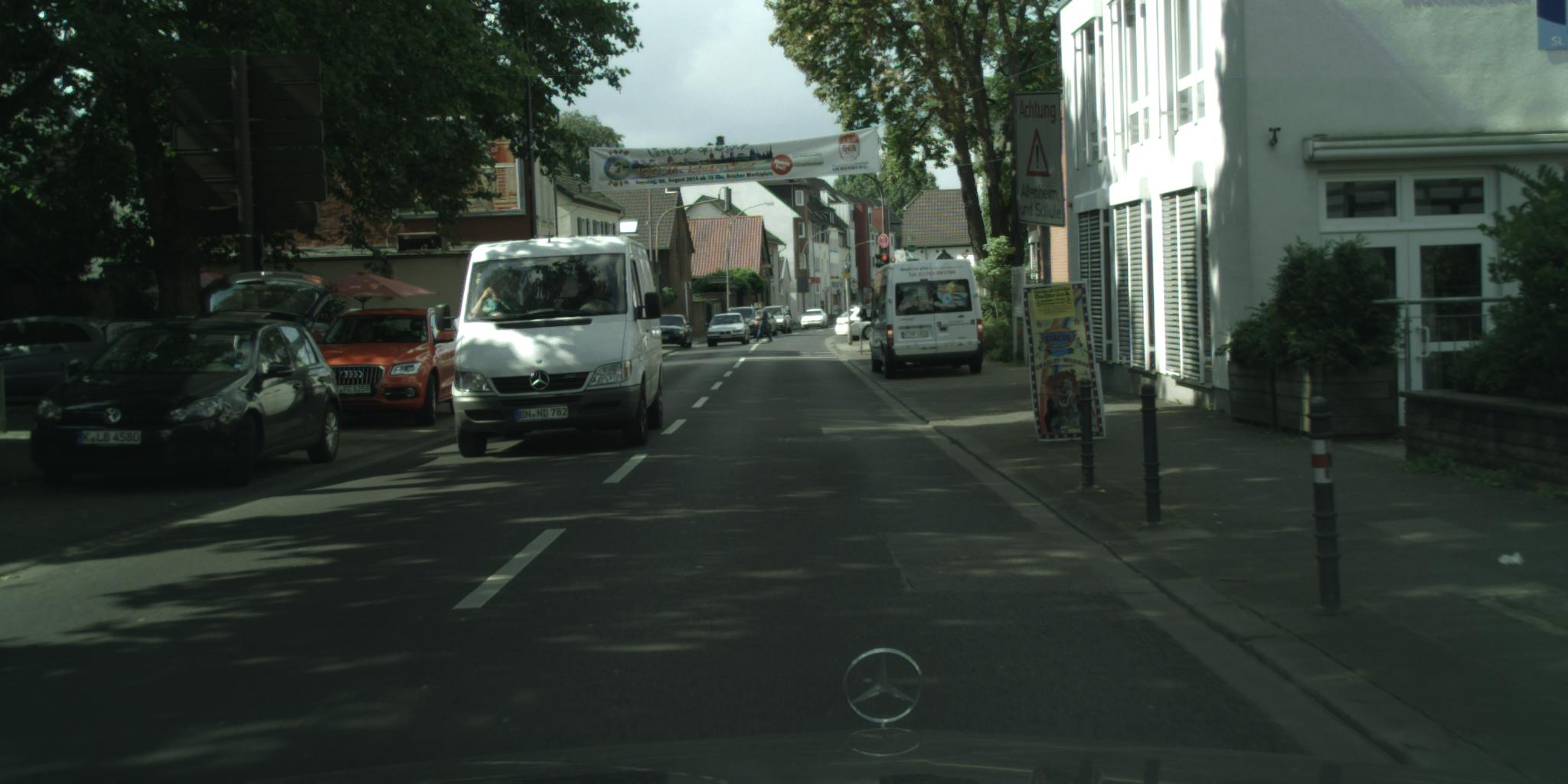}}
	\resizebox{0.245\columnwidth}{!}{%
		\includegraphics[scale=0.5]{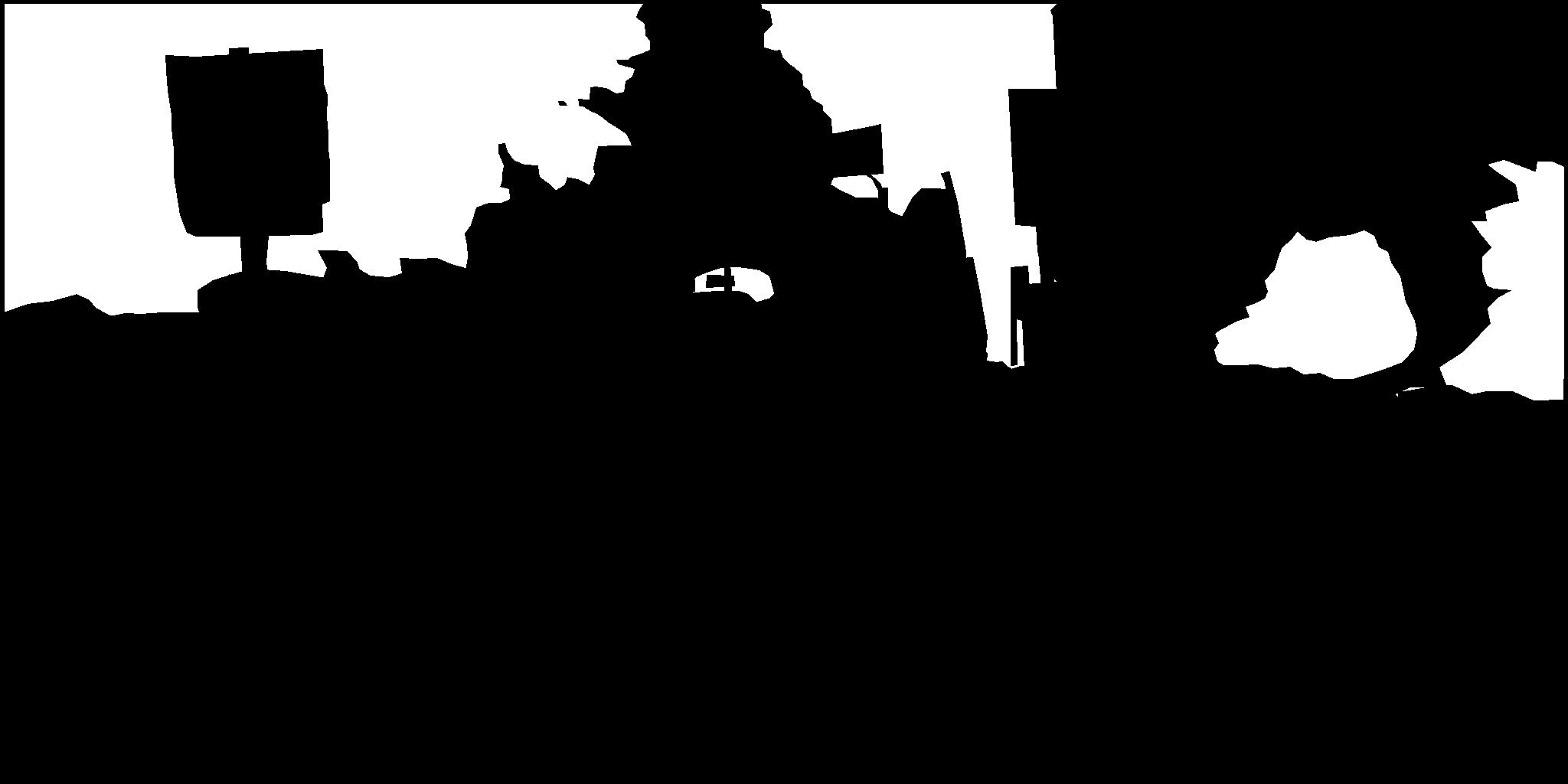}}
	\resizebox{0.245\columnwidth}{!}{%
		\includegraphics[scale=0.5]{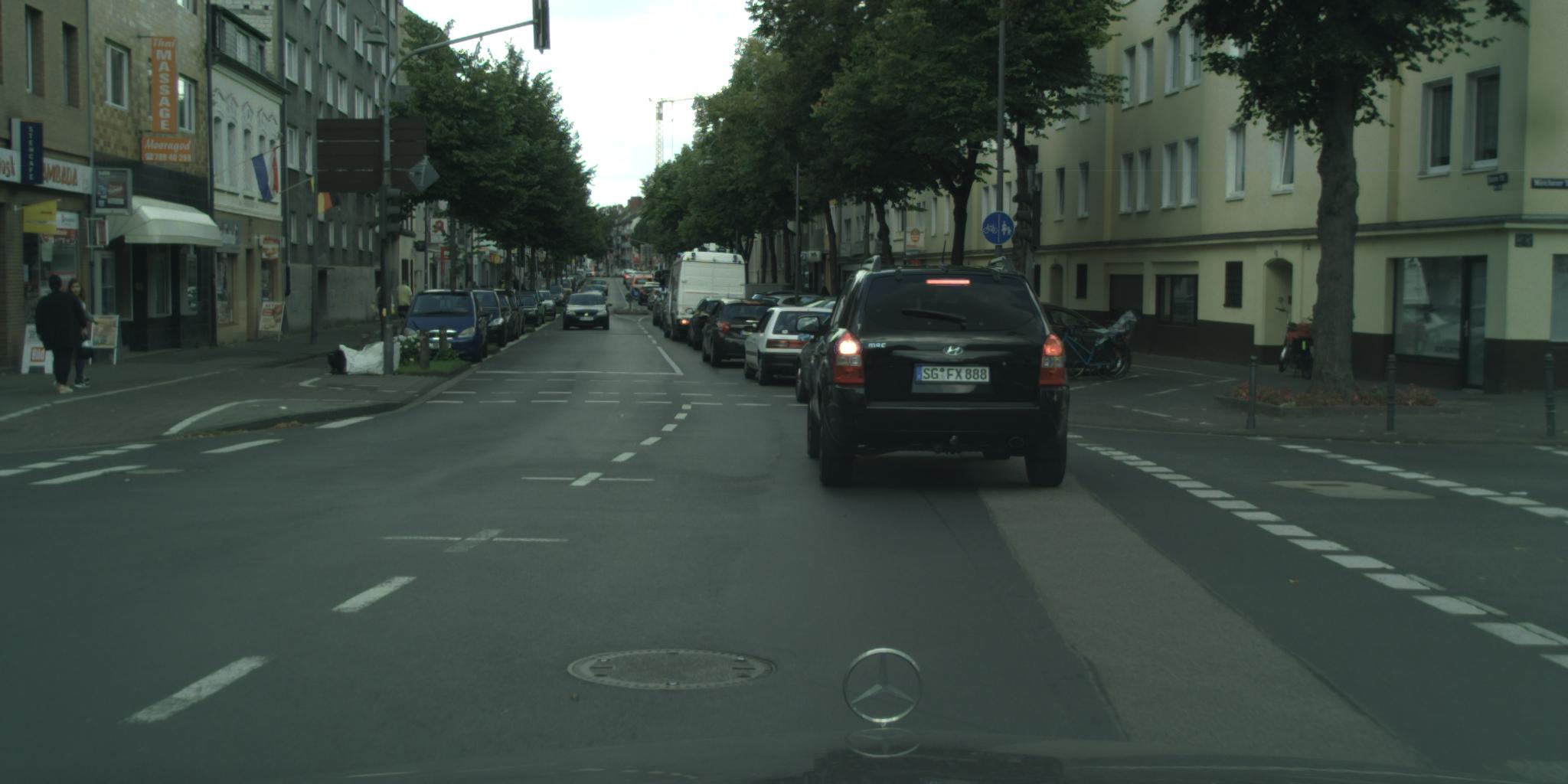}}
	\resizebox{0.245\columnwidth}{!}{%
		\includegraphics[scale=0.5]{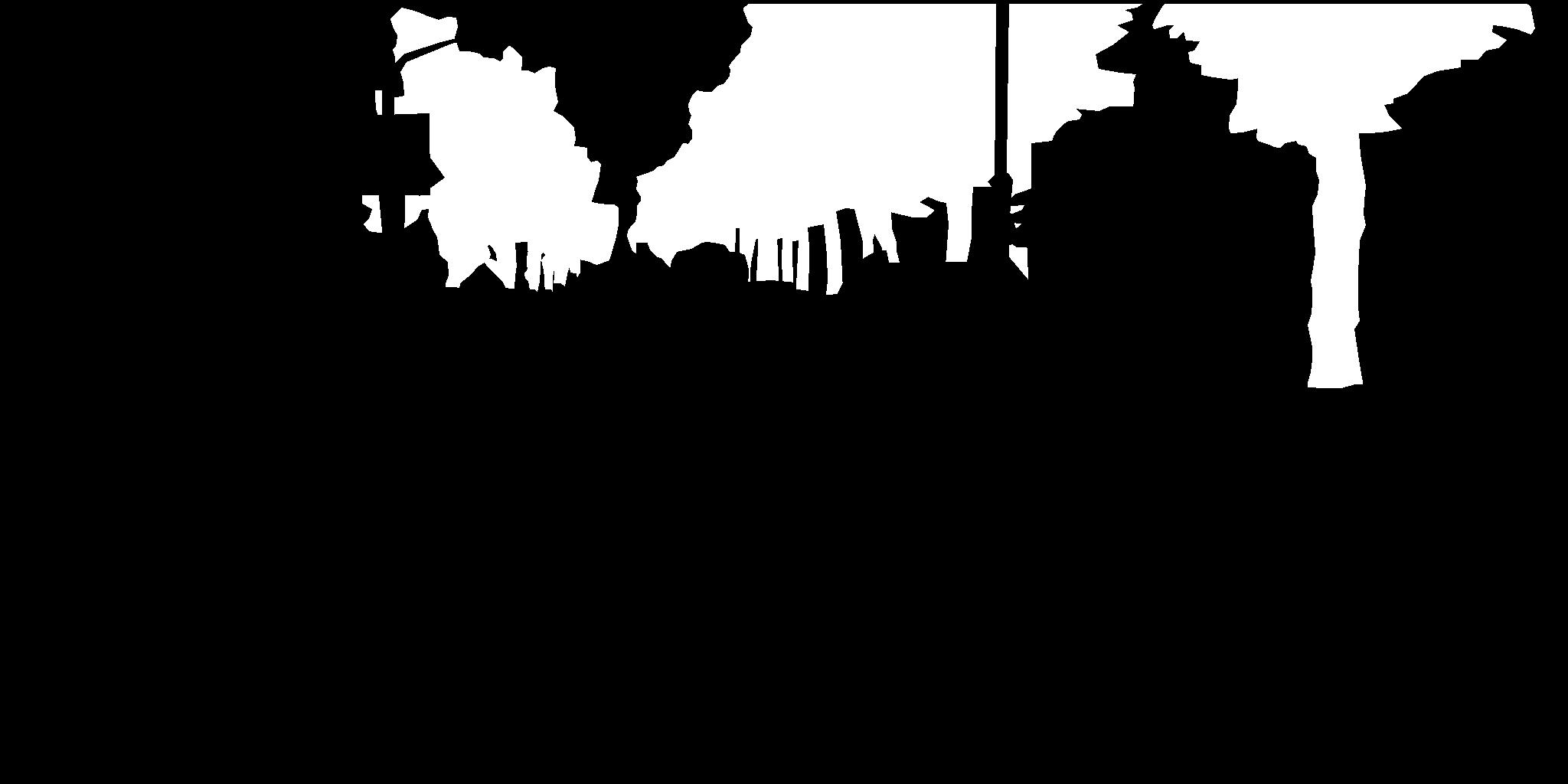}}
	\caption{Top: GSV image in Singapore and  S$\widetilde{\mbox{a}}$o Paulo, with their associated vegetation labels. Bottom: Two sample images from Cityscapes dataset and their associated vegetation labels}
\end{figure}

\subsection*{Appendix: Grad-CAM visualizations}

\begin{figure}[H]
	\centering
	\subfloat{
		\resizebox{0.5\columnwidth}{!}{
			\includegraphics[scale=0.2]{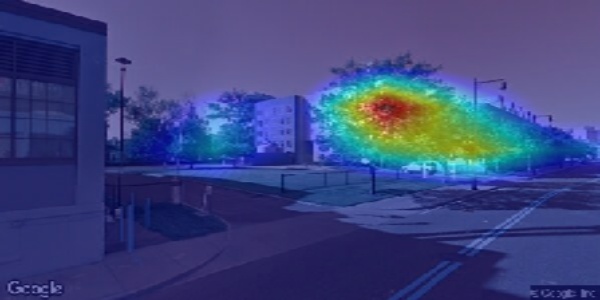}}}
	\subfloat{
		\resizebox{0.5\columnwidth}{!}{
			\includegraphics[scale=0.2]{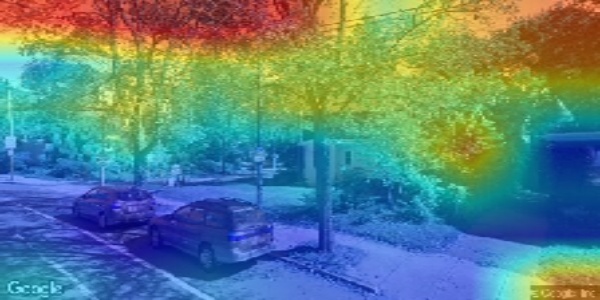}}}\\
	\subfloat{
		\resizebox{0.5\columnwidth}{!}{
			\includegraphics[scale=0.2]{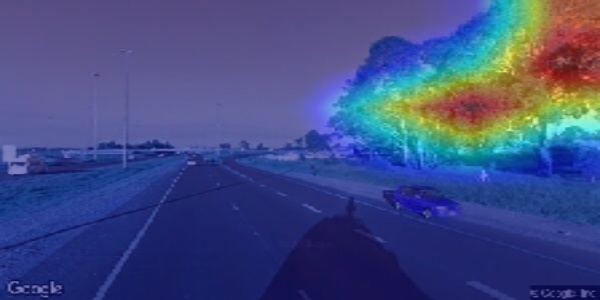}}}
	\subfloat{
		\resizebox{0.5\columnwidth}{!}{
			\includegraphics[scale=0.2]{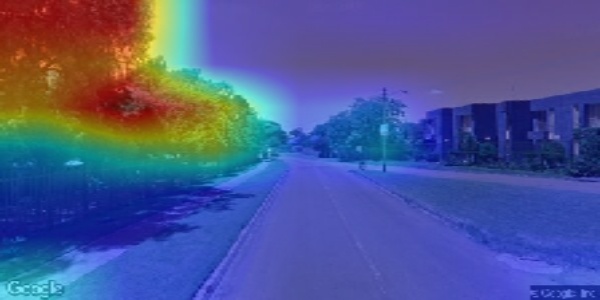}}}\\
	\subfloat{
		\resizebox{0.5\columnwidth}{!}{
			\includegraphics[scale=0.2]{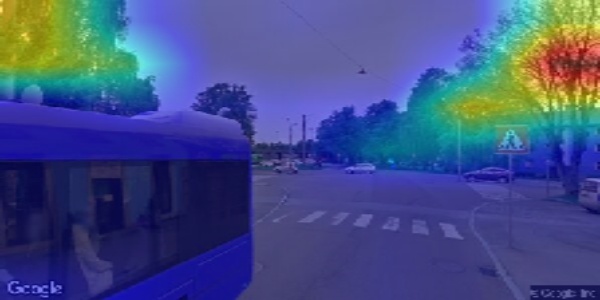}}}
	\subfloat{
		\resizebox{0.5\columnwidth}{!}{
			\includegraphics[scale=0.2]{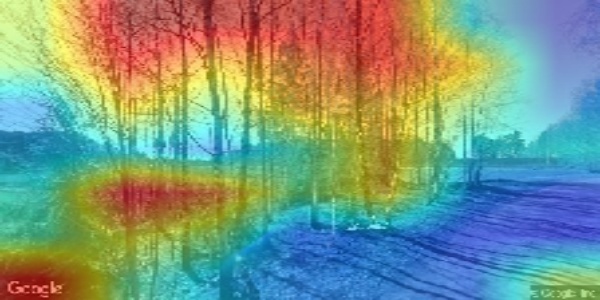}}}
\end{figure}
\newpage
\begin{figure}[H]
	\centering
\subfloat{
	\resizebox{0.5\columnwidth}{!}{
		\includegraphics[scale=0.2]{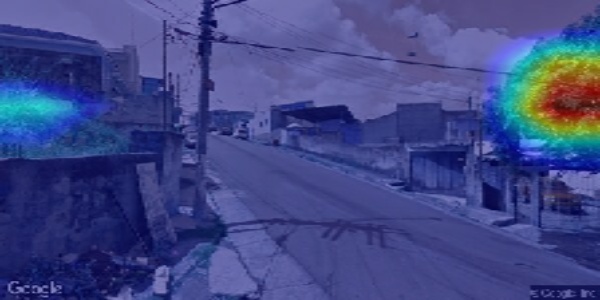}}}
\subfloat{
	\resizebox{0.5\columnwidth}{!}{
		\includegraphics[scale=0.2]{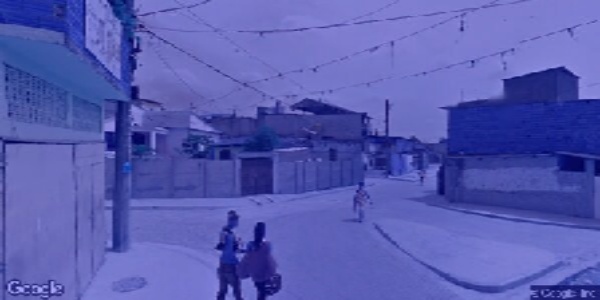}}}\\
\subfloat{
	\resizebox{0.5\columnwidth}{!}{
		\includegraphics[scale=0.2]{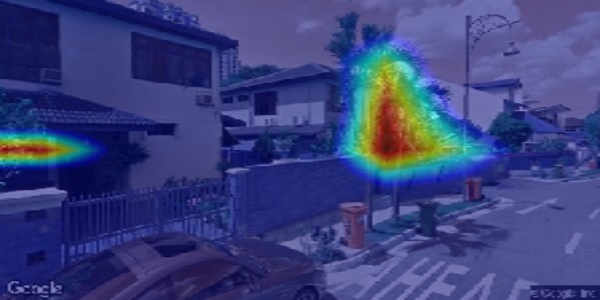}}}
\subfloat{
	\resizebox{0.5\columnwidth}{!}{
		\includegraphics[scale=0.2]{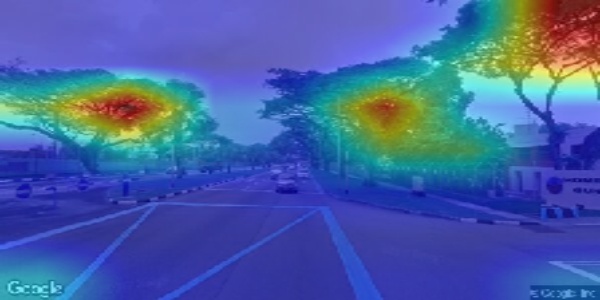}}}
	\caption{Results from applying Grad-CAM on our trained DCNN end-to-end modele to understand features learned in the last convolutional layer. 2 images from Cambridge, Johannesburg, Oslo, Sao Paulo, and Singapore are shown in order. Areas closer to red have a more positive contribution to a higher prediction of GVI than the contribution of areas closer to blue.}
\end{figure}
\end{document}